\documentclass[acmsmall,screen]{acmart}

\AtBeginDocument{%
  \providecommand\BibTeX{{%
    \normalfont B\kern-0.5em{\scshape i\kern-0.25em b}\kern-0.8em\TeX}}}

\setcopyright{rightsretained}
\copyrightyear{2024}
\acmYear{2024}
\acmDOI{XXXXXXX.XXXXXXX}

\acmISBN{978-1-4503-XXXX-X/18/06}

\usepackage{algorithm}
\usepackage{algpseudocode}
\usepackage{multirow}
\usepackage{array}

\begin{document}

\title{Benchmark Data Contamination of Large Language Models: A Survey}

\author{Cheng Xu}
\email{cheng.xu1@ucdconnect.ie}
\orcid{0000-0002-4675-9122}
\affiliation{%
  \institution{University College Dublin}
  \city{Dublin}
  \country{Ireland}
}

\author{Shuhao Guan}
\email{shuhao.guan@ucdconnect.ie}
\orcid{0009-0004-7892-5019}
\affiliation{%
  \institution{University College Dublin}
  \city{Dublin}
  \country{Ireland}}

\author{Derek Greene}
\email{derek.greene@ucd.ie}
\orcid{0000-0001-8065-5418}
\affiliation{%
  \institution{University College Dublin}
  \city{Dublin}
  \country{Ireland}}

\author{M-Tahar Kechadi}
\email{tahar.kechadi@ucd.ie}
\orcid{0000-0002-0176-6281}
\affiliation{%
  \institution{University College Dublin}
  \city{Dublin}
  \country{Ireland}}

\authorsaddresses{%
Authors' addresses: Cheng Xu, cheng.xu1@ucdconnect.ie; Shuhao Guan, shuhao.guan@ucdconnect.ie; Derek Greene, derek.greene@ucd.ie; M-Tahar Kechadi, tahar.kechadi@ucd.ie, School of Computer Science, University College Dublin, Belfield, Dublin 4, Ireland.
}

\renewcommand{\shortauthors}{Xu et al.}

\begin{abstract}
The rapid development of Large Language Models (LLMs) like GPT-4, Claude-3, and Gemini has transformed the field of natural language processing. However, it has also resulted in a significant issue known as Benchmark Data Contamination (BDC). This occurs when language models inadvertently incorporate evaluation benchmark information from their training data, leading to inaccurate or unreliable performance during the evaluation phase of the process. 
This paper reviews the complex challenge of BDC in LLM evaluation and explores alternative assessment methods to mitigate the risks associated with traditional benchmarks. The paper also examines challenges and future directions in mitigating BDC risks, highlighting the complexity of the issue and the need for innovative solutions to ensure the reliability of LLM evaluation in real-world applications.

\end{abstract}




\maketitle

\section{Introduction}
\label{sec:intro}

The field of natural language processing (NLP) has undergone a significant transformation in recent years, thanks to the rapid advancement of Large Language Models (LLMs) like GPT-4 \cite{openai2024gpt4}, Claude-3 \cite{anthropic2024introducing}, and Gemini \cite{geminiteam2023gemini}. These models, built on deep learning architectures such as Transformers \cite{vaswani2017attention}, have revolutionized various domains, including content generation, summarization, machine translation, and question-answering. By demonstrating remarkable capabilities in understanding and generating human-like text, they have gained widespread interest and acceptance in both academia and industry.

Amid the excitement surrounding the progress of LLMs, a critical issue has emerged: Benchmark Data Contamination (BDC). This refers to the phenomenon where language models incorporate information related to the evaluation benchmark from their training data, leading to skewed or unreliable performance during the evaluation phase. The challenge at hand involves both the evaluation process of LLMs and their privacy and security considerations \cite{gupta2023from,huang2022large,kandpal2022deduplicating,carlini2023extracting,carlini2021extracting}. While some studies see this phenomenon as beneficial \cite{blevins2022language} or do not consider it to be a problem \cite{cao2024concerned}, the majority of studies in the academic community agree that BDC poses significant challenges to the reliability and validity of LLM evaluations, undermining trust in their outputs and hindering their real-world applications \cite{lee2022deduplicating,sainz2023nlp,mcintosh2024inadequacies,zhou2023dont,jiang2024does,riddell2024quantifying}.

Traditional evaluation methodologies for LLMs often rely on benchmark datasets as gold standards for measuring model performance. Although these benchmarks are crucial for evaluating, validating, and comparing different models, they are not immune to the issue of BDC. With the rise of AI-generated content (AIGC), this issue is becoming more complex and difficult to detect. The datasets used for training and fine-tuning LLMs may contain benchmark-related information, such as metadata, label distributions, and contextual data, which can inadvertently impact the models' behavior and evaluation performance. Therefore, assessments based on traditional benchmarks may not accurately represent the true capabilities of LLMs and can lead to misguided conclusions about their performance.

In response to the widespread challenges around BDC, researchers have started to explore alternative assessment methods to reduce the risks associated with traditional benchmarks. Some promising approaches have been proposed, such as regenerating benchmark data \cite{xia2024leaderboard,zhu2024dyval,zhu2024dyvala}, which mitigates BDC by reconstructing the original benchmarks using LLMs, and benchmark-free evaluation \cite{li2024treeeval,yu2024freeeval,chiang2024chatbot}, which tries to avoid relying on predefined benchmarks altogether. These approaches aim to evaluate LLMs in a more flexible, adaptive, and reliable manner.

Along with the rapid development of LLMs, the issue of BDC has become increasingly important and observed in the research community. However, there is currently no comprehensive and systematic research that thoroughly discusses and defines this problem. This paper aims to fill this gap by providing a comprehensive survey on BDC in LLMs. In this survey, we define the BDC problem and organize the existing research into two main categories: Detection Techniques and Mitigation Strategies. The first category focuses on how to identify and detect BDC risks, while the second category focuses on mitigating the BDC problem in the current evaluation process of LLMs. By conducting this survey, we provide a comprehensive understanding of BDC in LLMs and offer insights into the detection and mitigation of this critical issue. 

This paper is organized as follows. Section \ref{sec:background} provides relevant background information about LLMs, and we define and discuss the BDC problem and provide some examples. Sections \ref{sec:detection} and \ref{sec:mitigation} comprehensively review existing methods for detecting BDC during the evaluation of LLMs and strategies for mitigating BDC risks, respectively. The detection methods are divided into two subcategories: Matching-based and Comparison-based methods. The mitigation strategies are further divided into three subcategories: Curating New Data, Refactoring Existing Data, and Benchmark-free Evaluation. Within each category, key approaches are discussed. Subsequently, Section \ref{sec:future} examines the challenges and future directions for mitigating BDC risks, acknowledging the inherent complexities and trade-offs involved in developing robust evaluation strategies for LLMs.

\section{Background}
\label{sec:background}

In this section, we provide an in-depth review of LLMs and BDC. Initially, we explore the current state of research on LLMs in Section \ref{sec:llms}. We then elucidate the concept of BDCs and provide a formal definition in Section \ref{sec:bdc}. In Section \ref{sec:sources}, we investigate the origins of BDC issues and their potential implications. Lastly, in Section \ref{sec:relatedTasks}, we identify a selection of critical tasks that are susceptible to the effects of BDC.

\subsection{Large Language Models}
\label{sec:llms}

An LLM is a language model notable for its ability to achieve general-purpose language understanding and generation. Such models have evolved significantly, leveraging advancements like Transformers and self-attention, which have enabled them to process longer sequences effectively. Examples of LLMs include GPT \cite{radford2019language,brown2020language,openai2024gpt4}, PaLM \cite{chowdhery2023palm}, and LLaMA \cite{touvron2023llama}. They serve as the backbone of many NLP applications, such as text generation, translation, question answering, and summarization. Research on LLMs has advanced in both academia and industry, and remarkable progress has been made with the launch of ChatGPT \cite{openai2022introducing}, which has attracted widespread attention. 

Earlier LLMs typically made use of encoder-decoder or encoder-only architectures, which are effective for various NLP applications. For instance, models like BERT (encoder-only) \cite{devlin2018bert} and T5 (encoder-decoder) \cite{raffel2020exploring} have shown strong performance in tasks such as text classification and machine translation. However, the most advanced and largest LLMs use a decoder-only transformer-based architecture \cite{roberts2024powerful}, while some recent variations are based on other architectures, such as recurrent neural network variants and Mamba \cite{gu2023mamba}. The later LLM architectures \cite{roberts2024powerful} can use unlabeled data for unsupervised pre-training and exhibit better generalization across different tasks compared to encoder-decoder and encoder-only architectures \cite{wang2022language}. There are four key hyper-parameters that characterize an LLM: the cost of pre-training, the model size, the dataset size, and the performance after pre-training.

The performance after (pre-)training is influenced by three other hyper-parameters: 1) model size, 2) training data size, and 3) cost of training. Consequently, some studies have explored the relationship between these hyper-parameters and the model's cross-entropy loss to understand how these factors affect model efficiency and performance. The KM scaling law \cite{kaplan2020scaling} focuses on the benefits of increasing model size and dataset size, while the Chinchilla scaling law \cite{hoffmann2022training} highlights the importance of balancing model size with the appropriate amount of data for optimal performance. Additional research also indicates that scaling can significantly enhance the capacity of LLMs \cite{brown2020language, chowdhery2023palm}. This improvement occurs because larger models can capture more complex patterns and relationships within the data. Additionally, increasing the amount of training data exposes the model to a broader range of information, further improving its generalization abilities and enabling it to handle diverse and challenging tasks more effectively.

LLMs primarily possess three basic abilities: language generation, knowledge utilization, and complex reasoning. Current mainstream LLMs perform language generation by proposing the next token based on previous tokens \cite{bengio2000neural}. LLMs can also generate specialized languages, such as programming code, via code synthesis \cite{gulwani2017program}. The ability of knowledge utilization refers to LLMs that can accomplish knowledge-intensive tasks through knowledge provided during pre-training and within prompts. This ability is primarily evaluated through question-answering (QA) tasks \cite{kadavath2022language} and knowledge graph completion tasks \cite{toutanova2015observed}. Complex reasoning refers to the ability to understand and use supporting evidence or logic to derive conclusions or make decisions \cite{huang2023towards}. This can be assessed through tasks, such as knowledge reasoning \cite{saikh2022scienceqa} and symbolic reasoning \cite{jie2023towards}.

LLMs also have some advanced abilities, such as interacting with external environments or user tools. Some studies have enabled LLMs to perform specific tasks like autonomous driving through external interfaces \cite{chen2023driving} or control characters in games to achieve specific goals \cite{wang2023advances}. When solving complex problems, LLMs can use external tools if deemed necessary, such as a search engine \cite{nakano2022webgpt}, image generation models \cite{betker2023improving}, and compilers \cite{gao2023pal}. Such tools are used to enhance the performance of LLMs in various applications. These capabilities stem from LLMs' proficiency in understanding context, generating relevant output, and interacting with other systems through well-defined interfaces, thereby enhancing their performance in various applications.

Emergent abilities manifest primarily in three ways: in-context learning \cite{dong2023survey}, instruction following \cite{sanh2022multitask}, and step-by-step reasoning \cite{prystawski2023think}. In-context learning abilities were first observed in GPT-3 \cite{brown2020language}. The model can adjust its responses based on the examples or instructions included within the same input prompt, and in-context learning does not require the model to undergo additional training or change its weights.

The instruction following ability enables the model to comprehend and execute tasks based on directives provided directly within the input prompt. For example, when provided with a prompt that includes specific instructions, such as ``summarize the following text" or ``translate the following sentences into French," the model can understand and act upon these instructions, leveraging its pre-existing knowledge base and generalization capabilities \cite{wei2022finetuned}.

Step-by-step reasoning refers to the ability of models to break down complex problems or queries into smaller steps, processing each one sequentially to arrive at a final answer. This is often accomplished through chain-of-thought (CoT) prompting strategy \cite{wei2022chain}, in this approach, the model sequentially processes each step, building upon prior steps to construct a comprehensive answer, and research has shown that chain-of-thought prompts can bring performance gains \cite{wei2022chain}.

The abilities mentioned above enable LLMs to exhibit strong performance but also face several issues. A frequently-reported problem is the generation of so-called hallucinations \cite{bang2023multitask}, where LLMs generate text that superficially appears to be correct but is actually inaccurate. This problem is difficult to resolve completely, although it can be mitigated through alignment tuning strategies \cite{lee2009advances}. While LLMs have learned general language patterns, they underperform in specialized domains, such as medicine or engineering. This may be related to catastrophic forgetting \cite{kemker2018measuring} or a scarcity of relevant training data. Furthermore, enabling LLMs to quickly learn the latest knowledge by updating weights remains an unresolved challenge \cite{yao2023editing}.

\subsection{Benchmark Data Contamination}
\label{sec:bdc}

Benchmark Data Contamination refers to a critical issue encountered during the training and evaluation of LLMs. It arises when an LLM inadvertently encounters test data (or benchmark data) during its training and fine-tuning process. This exposure can significantly impact the model's performance scores, leading to inflated results that do not accurately reflect its true capabilities. We formally define this phenomenon as follows.

\begin{definition}[Benchmark Data Contamination]
Exposure of a large language model to benchmark data during the training process leads to distorted evaluation results.
\end{definition}

\noindent Depending on the severity of the contamination, we categorize the BDC problem into four types:
\begin{enumerate}
    \item \textbf{Semantic Level}: Exposure of identical and/or derivative content of the benchmark. Typically, the content pertains to the same topic or comes from the same source as the benchmark. This form of contamination introduces biases related to specific topics, affecting the model's generalization capabilities.
    \item \textbf{Information Level}: Exposure to benchmark-related information leads to models with tendencies and biases during evaluation. Information such as metadata, time distributions, label distributions, and external reviews of the benchmark can inadvertently influence the model's evaluation process.
    \item \textbf{Data Level}: The benchmark data exposure excludes labels. Examples include the data content of the test set and data sequences without associated labels. Data-level contamination affects the model's understanding of the underlying patterns and relationships within the data.
    \item \textbf{Label Level}: The complete exposure of benchmark data, including labels. When labels are made available during training, the model may directly memorize them, leading to overfitting and compromised generalization.
\end{enumerate}
As we move from the semantic level to the label level, the severity of BDC increases, posing greater challenges to the evaluation of models. However, the complexity of detecting and preventing contamination inversely correlates with the proximity to full exposure to the benchmark. While complete exposure at the label level facilitates relatively straightforward detection and prevention measures, the more abstract and intricate nature of contamination at the semantic and information levels renders it inherently more challenging to detect and mitigate.

\subsection{Sources and Impact}
\label{sec:sources}

The training processes of LLMs can be divided into two main types. Pre-training refers to the process of training language models on a large-scale corpus with the aim of equipping LLMs with general-purpose language comprehension, where the trained model is called Pre-trained Language Models (PLMs) \cite{qiu2020pre}. In contrast, fine-tuning refers to the targeted continuation of training for various downstream natural language processing tasks, in order to make the LLMs better able to carry out the downstream tasks.

The BDC problem stems from the inherent complexity and diversity of the pre-training data used to train LLMs in NLP tasks. Excluding deliberate human attacks, one of the primary sources of BDC is the composition of large-scale pre-training datasets themselves. These datasets are often compiled from a wide range of sources, including news articles, online forums, social media posts, and other publicly available text data. While this diversity is essential for training robust and generalizable models, it also introduces the risk of unexpected exposure to test data during the model training process. Such a risk, on the other hand, is much less in the fine-tuning process, which generally uses relatively small datasets for targeted training, and is a much more controlled process with more predictable results.

\begin{table}[t]
\centering
\caption{Key metrics used in different aspects of automatic evaluation of LLMs.}
\label{tab:key_metrics}
\begin{tabular}{l|l}
\hline
\textbf{Criteria} & \textbf{Metrics}                  \\ \hline
Accuracy                & Exact match, Quasi-exact match, F1 score, ROUGE score \cite{zeng2023evaluating} \\
Calibrations            & Expected calibration error \cite{guo2017calibration}, Area under the curve \cite{geifman2017selective}  \\
Fairness                & Demographic parity difference \cite{zemel2013learning}, Equalized odds difference \cite{hardt2016equality} \\
Robustness              & Attack success rate \cite{wang2022adversarial}, Performance drop rate \cite{zhu2023promptbench}       \\ \hline
\end{tabular}
\end{table}

If BDC is introduced during the training phase, we need to consider how this might impact on the evaluation process of LLMs. Such evaluations typically involve three methods: traditional benchmark testing, automatic evaluation, and human evaluation. Each method has its own metrics and processes, which can be significantly affected by BDC.

In traditional benchmark testing, the model's performance is assessed by training or fine-tuned on a training set and then testing it on a separate test set. However, the presence of BDC can lead to overestimated performance metrics, as the model might inadvertently ``learn" from test data that was leaked into the training set. This compromises the integrity of the evaluation, making it difficult to gauge the model's true capabilities.

Automatic evaluation uses algorithms and pre-defined metrics to assess LLMs, reducing the need for human labor and enabling faster, more standardized assessments. Key aspects of automatic evaluation include: Accuracy, Calibration, Fairness, Robustness, which are presented in Table \ref{tab:key_metrics}. Accuracy measures the correctness of the model's outputs against a ground truth. Calibration assesses how well the model's confidence aligns with its accuracy. Fairness evaluates the bias in the model's outputs across different demographic groups. Robustness tests the model's resilience to adversarial attacks and perturbations. This method is also currently recognized as the most promising evaluation strategy. For example, \citet{jain2023bring} introduced a self-supervised method to streamline the evaluation of models in real-world deployments by eliminating the need for labeling new data. Additionally, in some studies, LLMs have been used as judge models to evaluate the performance of other LLMs \cite{wang2024pandalm, huang2024empirical}. Automatic evaluation can mitigate some effects of BDC by employing self-supervised methods that reduce reliance on labeled data, but this approach cannot completely eliminate the risk of contamination, for example, the LLM used for automatic evaluation may also suffers from the BDC problem. The effects on the evaluation metrics are as follows:
\begin{itemize}
    \item \textbf{Accuracy}: Directly impacted as the model might have already seen the test data, leading to inflated accuracy scores.
    \item \textbf{Calibration}: Can be skewed if the model's confidence scores are based on contaminated data, giving a false sense of reliability.
    \item \textbf{Fairness}: Potentially affected if the contaminated data introduces or reinforces biases that the model learns and propagates.
    \item \textbf{Robustness}: Compromised because the model's resilience to unseen data and adversarial conditions is not accurately tested with contaminated data.
\end{itemize}

Human evaluation is considered to be the most rigorous and nuanced method, where one or more human judges assess the model's performance on various criteria \cite{novikova2017we}. For example, the Chatbot Arena created by \citet{chiang2024chatbot} has gathered numerous human votes. Human evaluations often follow principles like the 3H rule (Helpfulness, Honesty, Harmlessness \cite{askell2021general}) or specific criteria \citet{chang2023survey} such as accuracy \cite{singhal2023large}, relevance \cite{zhong2022unified}, fluency \cite{van2019best}, transparency \cite{wu2022ai}, safety \cite{ji2024beavertails}, and human alignment \cite{ouyang2022training}. While human evaluation can provide a more realistic assessment of an LLM's performance, it is also susceptible to biases from evaluators' backgrounds and experiences. BDC poses a unique challenge to human evaluation. While human judges might be able to recognize and mitigate some effects of data contamination, their subjective judgments can still be influenced by familiarity with the data. Moreover, if the human evaluators themselves are biased by prior exposure to the contaminated data, their assessments might not fully reflect the model's true performance on genuinely novel inputs.

In conclusion, the sources and impact of BDC are critical considerations in the training and evaluation of LLMs. BDC arises primarily from the diverse and extensive pre-training datasets, potentially leading to overestimated performance metrics in traditional benchmark testing and skewed results in both automatic and human evaluations. While automatic evaluation offers a promising approach to mitigate some BDC effects through standardized and self-supervised methods, it cannot fully eliminate the risk. Human evaluation, despite being the most nuanced and rigorous method, is also vulnerable to biases introduced by BDC. Therefore, addressing BDC is essential for ensuring the integrity and reliability of LLM assessments across all evaluation methods.

\subsection{Related Tasks}
\label{sec:relatedTasks}

To gain a clearer understanding of the prevalence of the BDC problem in NLP tasks, we systematically select seven common LLM tasks and delineate specific instances where BDC vulnerability can frequently manifest:
\begin{itemize}
    \item \textbf{Code Generation} \cite{jain2024livecodebench,xia2024leaderboard,guo2024deepseekcoder,rozière2024code,chen2021evaluating,majdinasab2024trained}: In code generation tasks, large-scale pre-training data may include code snippets and corresponding programming ideas from online forums or repositories about the benchmarks, which may lead to a high risk of contamination. For example, if content related to the benchmark is included in the pre-training process, while it is not difficult to mitigate BDC by directly filtering answers that match the test set, excluding semantic-level related content, such as problem-solving tutorials, is challenging. Thus the model may still be at risk of contamination, leading to distorted evaluation results.
    \item \textbf{Machine Translation} \cite{hendy2023good,zhang2023prompting,vilar2023prompting,zhu2023multilingual,moslem2023adaptive,bawden2023investigating,jiao2023chatgpt}: In machine translation tasks, benchmarks are often composed of translated texts from various common sources, which can naturally lead to contamination. For example, news articles or official announcements are usually available in a variety of languages, and testing a model using benchmarks that cover that topic can make it biased towards certain topics or narrative structures, leading to an inaccurate assessment of the model's translation capabilities.
    \item \textbf{Question Answering} \cite{lewis2021question,pampari2018emrqa,singhal2023expertlevel,robinson2023leveraging,ouyang2022training,talmor2019commonsenseqa,khashabi2020unifiedqa,liu2023webglm,dunn2017searchqa,rogers2023qa}: Benchmarks for QA tasks usually contain question and answer pairs. However, if relevant content, like discussions on Github Issues\footnote{\url{https://docs.github.com/en/issues/tracking-your-work-with-issues/about-issues}}, is introduced during the pre-training process, models trained on such data may have difficulty recognizing correct answers, resulting in inflated performance scores that do not reflect the true functionality of the model.
    \item \textbf{Sentiment Analysis} \cite{Aggarwal2018,medhat2014sentiment,zhang2018deep,dai2021syntax,zhang2023sentiment,yadav2020sentiment,wang2024chatgpt,zhang2023a,zhang2021towards,althubaity2023evaluating}: In sentiment analysis tasks, common benchmarks consist of text samples labeled with sentiment under a certain topic. If the contextual information of the topic is contained in the pre-training data, it gives the model subjective biases and tendencies, which can lead to distorted results in sentiment prediction evaluation. For example, in the sentiment prediction task about COVID, if the background information of this topic is present in the pre-training information, it will let the model know in advance that it is a worldwide infection, resulting in a pre-determined distribution of predominantly negative sentiment labels from the model.
    \item \textbf{Named Entity Recognition} \cite{wang2023gptner,elangovan2021memorization,lucy2021gender,liang2021towards,devassimonmanela2021stereotype,nangia2020crows,blodgett2020language,sang2003introduction,isaak2023pronounflow,stammbach2022heroes}: Benchmark for the NER task consist of text annotated with named entities (e.g., names of people, organizations, and locations). However, if the pre-training material contains content related to these entities, it can let the model learn about the prior background knowledge, which will lead to a distorted evaluation of the entity recognition performance.
    \item \textbf{Fake News Detection} \cite{pelrine2023towards,blevins2022language,elazar2023measuring,wang2017liar,shu2020fakenewsnet,xu2023fuzzy,Hu2024bad,aman2024large,zhou2020a,wang2022archival}: Articles and comments associated with news events that constitute a benchmark for the fake news detection task might be used as pre-training data, leading to a risk of BDC. An event is usually covered by more than one media outlet, and different media outlets may have different positions and languages. Such large-scale relevant information for model training may lead to multiple BDC problems, from the semantic level to the label level.
    \item \textbf{Text Reconstruction} \cite{chang2023speak,karsdorp2019cultural,spennemann2023chatgpt,moskvichev2023narrativexl,onishi2016large,ferguson2020iirc,xie2018large,zheng2019chid}: The benchmark for text reconstruction tasks generally consists of incomplete or fragmented text passages. If the pre-training dataset contains complete benchmark texts, e.g., the original text in an antiquarian book restoration task is already present in the pre-training data, this can lead to serious distortions in the evaluation results.
\end{itemize}

From the above, we see that BDC poses a significant challenge in the training and evaluation of LLMs in many different contexts. In each case, contamination can potentially lead to distorted performance scores that do not accurately reflect the model's true capabilities. The severity of BDC, as categorized into Semantic Level, Information Level, Data Level, and Label Level, increases as we move closer to full exposure of the benchmark data. The complexity of detecting and mitigating BDC inversely correlates with the severity of exposure, making it a challenging problem to address. The primary sources of BDC are the large-scale pre-training datasets used in training LLMs, which due to their diversity and complexity, can inadvertently introduce the risk of BDC. Highlighting the potential scenarios of BDC occurrence across seven prevalent LLM tasks underscores the critical necessity of addressing this issue for precise model evaluation and performance enhancement in the domain of NLP.

\begin{table*}
  \caption{An overview of the main methods for detecting data contamination, listing their categories, together with a short description and representative references.}
  \label{tab:detection}
  \begin{tabular}{l|m{2cm}|m{5cm}|m{2cm}}
    \toprule
    \textbf{Type} & \textbf{Method} & \textbf{Short Description} & \textbf{References} \\
    \midrule

    \multirow{6}{*}{Matching-based} & Dataset\newline Inspection & Detect overlapping content between pre-training and evaluation datasets. & \cite{zhou2023dont,li2024task,balloccu2024leak,ippolito2023preventing,elangovan2021memorization,openai2024gpt4,brown2020language} \\ \cline{2-4}

    & Membership Inference & Make the model generate content based on test prompts for checking the inclusion of pre-training data. & \cite{golchin2024time,li2024open,li2024task,ranaldi2024investigating,deng2023investigating,golchin2024data,chang2023speak,shokri2017membership,shi2024detecting,duarte2024decop,ishihara2023training,koh2017understanding} \\ \cline{2-4}
    
    & Example\newline Generation & Make the model generate task-relevant examples for overlap checking. & \cite{li2024task} \\ \hline
    
    \multirow{7}{*}{Comparison-based} & Content\newline Comparison & Compare the difference between model-generated content or with evaluation dataset, e.g. similarity, distribution, perplexity. & \cite{golchin2024time,dong2024generalization,riddell2024quantifying,deng2023investigating,li2023estimating,magar2022data,shi2024detecting,lee2023platypus} \\ \cline{2-4}

    & Sequential Analysis & Assess the sequence alignment of model-generated content with the evaluation dataset. & \cite{oren2024proving,kandpal2022deduplicating} \\  \cline{2-4}
    
    & Chronological Analysis & Model performance is gauged on a dated dataset, assessing the impact of varying training data collection times. & \cite{li2024task,ranaldi2024investigating,huang2023competitionlevel,roberts2023data,yang2023rethinking,wu2024mrke,bordt2024elephants} \\
    
  \bottomrule
\end{tabular}
\end{table*}

\section{BDC Detection Techniques}
\label{sec:detection}

Efficient identification of BDC in evaluation benchmarks represents a fundamental aspect in ensuring the reliability and integrity of LLMs, while also providing the basis for developing effective mitigation strategies. This section provides a comprehensive review of literature related to the detection of BDC. We categorize the methodologies into two distinct strategies: matching-based and comparison-based, discussed in Sections \ref{sec:matching} and \ref{sec:comparison}, respectively. Note that certain investigations incorporate elements from both strategies. Such instances are allocated to the category deemed more comprehensive or preferred by the authors in question. All reviewed work on BDC detection is summarized in Table \ref{tab:detection}.

\subsection{Matching-based Methods}
\label{sec:matching}

These methods focus on detecting BDC by examining the overlap and inclusion of pre-training data in the evaluation datasets. This typically involves \textbf{dataset inspection}, \textbf{membership inference}, and \textbf{example generation}. We now discuss seven representative works in this area.

The predominant decontamination technique in NLP is n-gram overlap. Specifically, the work by \citet{brown2020language} associated with GPT-3 defines a 13-gram overlap as being indicative of contamination. In contrast, the more recent GPT-4 model \cite{openai2024gpt4} identifies a 50-character overlap as a contamination signal. N-gram overlap detection is favored for its simplicity and computational efficiency. However, it is essential to recognize that this approach may yield a higher false negative rate when dealing with subtle differences in text segments.

\citet{li2024task} investigated the phenomenon of task contamination in LLMs, like GPT-3, which may compromise their zero-shot and few-shot learning capabilities. The study revealed that LLMs perform better on datasets released before their training data creation date, suggesting the presence of contamination. The authors employed four methods: \textit{training data inspection} (Find examples of task training by inspecting the training data), \textit{task example extraction} (Extraction of task data from existing models), \textit{membership inference} (Check that the model-generated content of the input instance is identical to the original dataset.), and \textit{chronological analysis} (Measuring performance on datasets with known release dates and checking the evidence of contamination using chronological evidence) to provide evidence of this contamination. The paper also finds that for classification tasks without task contamination, LLMs show no significant improvement over simple majority baselines. 

\citet{ranaldi2024investigating} introduced another method for detecting BDC in GPT models. They assessed GPT-3.5's performance using the well-known Spider Dataset \cite{yu2018spider} and a novel dataset called Termite. Additionally, they employed an adversarial table disconnection (ATD) approach, which complicates Text-to-SQL tasks by removing structural pieces of information from the database. This method allowed them to analyze GPT-3.5's efficacy on databases with modified information and assess the impact of BDC on the model's performance.

\begin{figure}[h]
  \centering
  \includegraphics[width=130mm]{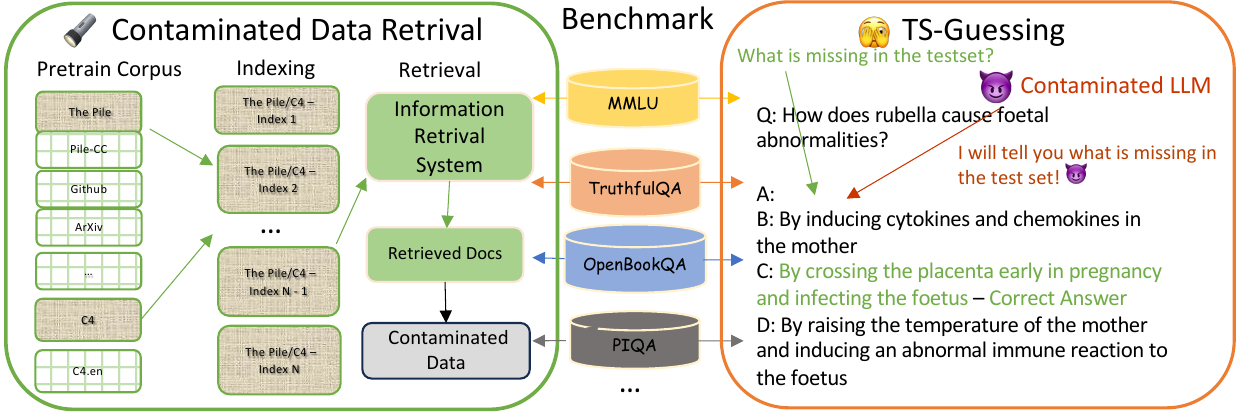}
  \caption{An illustration of the method developed by \citet{deng2023investigating} for identifying BDC in modern benchmarks. The figure on the left shows the workflow of an information retrieval system, which aims to detect potentially contaminated data within a benchmark by utilizing a pre-trained corpus. The figure on the right introduces TS-Guessing, an approach for detecting potential contamination. This technique involves concealing parts of the information in the test set and prompting LLMs to infer the missing elements. If the LLMs can accurately predict the same missing option as the one in the test set, it raises the suspicion that they may have encountered the benchmark data during their training.}
  \label{fig:deng2023investigating}
\end{figure}

Similarly, as shown in Figure \ref{fig:deng2023investigating}, \citet{deng2023investigating} proposed two novel methods to detect potential overlaps between evaluation benchmarks and pre-training corpora, tailored for both open-source and proprietary LLMs. They introduced a retrieval-based system and a Testset Slot Guessing (TS-Guessing) protocol, which involves masking incorrect answers in multiple-choice questions and prompting the model to fill in the gaps. Their findings indicated that commercial LLMs, including ChatGPT and GPT-4, can guess missing options in benchmark tests with a high level of accuracy.

\citet{golchin2024data} presented another novel approach, the Data Contamination Quiz (DCQ), to detect and quantify BDC in LLMs. The DCQ is a series of multiple-choice questions with three perturbed versions of each dataset instance, including only word-level changes. The LLM's ability to identify the original instance among the perturbed ones indicates potential exposure to the data during pre-training. Tested on GPT-3.5/4, the DCQ demonstrated higher contamination levels than other methods and effectively bypassed safety filters designed to prevent the generation of copyrighted content.

\citet{golchin2024time} presented a technique that combines instance-level identification using guided instruction prompts with partition-level assessment through overlap score comparison and classifier-based detection. The approach achieved high accuracy rates, between 92\% and 100\%, across seven datasets. The authors also found specific cases of contamination in popular datasets, such as AG News, WNLI, and XSum, when tested with GPT-4.

\citet{li2024open} presented a comprehensive report on BDC across over 15 LLMs and six multiple-choice QA benchmarks. They introduced an open-source pipeline to conduct contamination analysis on customized data and models. Their research uncovered varying degrees of contamination, ranging from 1\% to 45\%, and demonstrated that contamination does not always correlate with improved model performance. Interestingly, larger models may benefit more from contaminated test sets than smaller ones, with significant accuracy boosts observed on certain benchmarks.

Similar findings have been made in the field of historical book archaeology, \citet{chang2023speak} employed a membership inference query method called \textit{name cloze} to deduce the books known to these models. By removing the names from the work and retaining only the contextual information, the model is made to fill in the names in the form of cloze test. Based on the context alone, it should be almost impossible for the model to fill in the correct name, which requires not knowledge of English but specific knowledge of the work. The authors use this detection method in order to detect the severity of the BDC problem. Their findings reveal that the degree of memorization correlates with the frequency of book passages appearing on the web. This memorization affects the validity of cultural analytics assessments, as models perform better on memorized books than non-memorized ones in downstream tasks.

In this section, we have reviewed matching-based strategies employed to identify BDC within benchmarks. These strategies aim to uncover direct evidence, such as identifying matches between training data or LLMs-generated content, and the evaluation dataset. Common methods include inspecting the training set, generating LLMs' content related to the evaluation dataset for membership inference, and adjusting prompts to align with the evaluation dataset's content. This strategy offers the advantage of intuitively detecting BDC, enabling the development of mitigation strategies based on these detection techniques. However, it does have drawbacks. For instance, accessing the training set for data inspection is often impractical for commercially proprietary LLMs like GPT-4 \cite{openai2024gpt4} and Claude-3 \cite{anthropic2024introducing}. Additionally, the computational requirements and expertise needed for data matching pose challenges to widespread adoption, especially in resource-constrained settings. Notably, there are studies questioning the effectiveness of this detection scheme. \citet{yang2023rethinking}, \citet{ippolito2023preventing}, and \citet{jiang2024does} criticize the use of string matching methods like n-gram overlap for decontamination, demonstrating that simple test data variations such as paraphrasing can bypass these measures. They show that a 13B model can overfit a test benchmark and achieve high performance comparable to GPT-4 when such test data variations are not eliminated. Similar findings were reported by \citet{dekoninck2024evading}, who developed a technique called Evasive Augmentation Learning (EAL). This method involves rephrasing benchmark samples during the fine-tuning stage to evade detection by current contamination detection methods. They categorized model providers and contamination detection methods, uncovering vulnerabilities that EAL exploits. The technique proved highly effective, allowing significant improvements in benchmark performance (up to 15\%) while remaining undetected by existing contamination detection methods.

\subsection{Comparison-based  Methods}
\label{sec:comparison}

Another strategy to detecting BDC involves comparing the performance of model-generation on evaluation datasets. Common examples such as comparing the similarity  Common methods include comparing the similarity \cite{magar2022data,lee2023platypus}, distribution \cite{dong2024generalization}, perplexity \cite{li2023estimating}, and generation order \cite{oren2024proving} of the generated content with that of the evaluated dataset. Additionally, comparing the performance differences of LLMs on datasets across different time periods can serve as a comparison-based method for detecting BDC \cite{huang2023competitionlevel,roberts2023data}. We have identified six representative works that adopt this approach, which we have categorized into three subcategories: \textbf{content comparison}, \textbf{sequential analysis}, and \textbf{chronological analysis}. We discuss these below.

\citet{magar2022data} presented a method to detect contaminated data in downstream tasks, they tested pre-training BERT models on corpora that include Wikipedia and labeled downstream datasets, then fine-tuning them on relevant tasks, and then detected BDC by comparing the performance of model-generated content from "seen" and "unseen" evaluation datasets. Their experiments reveal that while some models do exploit contaminated data, others merely memorize them without exploitation. The study shows that the level of memorization and exploitation is influenced by factors such as the number of data duplications and model size. 

\citet{dong2024generalization} focused on the distribution of generated content, they proposed a novel method, CDD (Contamination Detection via output Distribution), which uses the output distribution of LLMs to detect BDC. They also introduce TED (Trustworthy Evaluation via output Distribution), which corrects the output distribution to mitigate the effects of contamination. Through extensive experiments, they demonstrate that CDD can significantly improve contamination detection over existing methods and TED can reduce performance inflation due to contamination. The paper also presents two new benchmarks, DetCon and ComiEval, for assessing BDC and mitigation methods. Their findings reveal that popular models like ChatGPT are susceptible to BDC, emphasizing the need for more reliable evaluation methods.

Different from the distribution, \citet{li2023estimating} proposed a novel method to detect BDC in language model evaluation without requiring access to the full training set. The technique uses perplexity to measure the extent of contamination, providing evidence of significant memorization in recent foundation models across various benchmarks. The study reveals that while reading comprehension and summarisation benchmarks show signs of contamination, multiple-choice benchmarks appear less affected. This method allows for a more accessible and less computationally intensive way to audit language models for contamination, ensuring more reliable evaluations.

An alternative interesting perspective is to focus on the order of content generated by LLMs. \citet{oren2024proving} presented a method to detect test set contamination in language models without needing access to the model's pre-training data or weights. The authors used a statistical test to identify contamination by comparing the likelihood of a benchmark dataset's canonical ordering against a shuffled version. Their findings suggest that it is possible to provide provable guarantees of test set contamination, which is significant for models trained on vast internet data. They successfully applied this test to audit popular language models and found minimal evidence of widespread contamination.

It is also possible to examine differences in performance based on data over time. \citet{huang2023competitionlevel} explored the reasoning capabilities of LLMs by using competition-level programming problems from Codeforces\footnote{\url{https://codeforces.com/}}. The study provided a comprehensive evaluation of GPT-4's performance on these problems, considering aspects such as release time, difficulty, and error types. The results showed a significant decline in GPT-4's performance on problems released after September 2021, indicating potential BDC and the challenges LLMs face with complex reasoning tasks. Despite exploring various approaches, like fine-tuning and Chain-of-Thought prompting, none consistently mitigated these challenges. The study underscores the value of competition-level problems as a resource for assessing LLMs' reasoning abilities and encourages the development of models with stronger reasoning skills and better generalization, and the study by \citet{yang2023rethinking} had a similar suggested scheme.

\begin{table*}[t]
  \caption{An overview of some of the main strategies for mitigating data contamination. We list their categories, as well as the corresponding short description and some representative references.}
  \label{tab:mitigation}
  \begin{tabular}{l|m{2cm}|m{5cm}|m{2cm}}
    \toprule
    \textbf{Type} & \textbf{Method} & \textbf{Short Description} & \textbf{References} \\
    \midrule

    \multirow{4}{*}{Data Curation} & Private\newline Benchmark & Isolating newly collected evaluation data to prevent its inclusion in pre-training datasets of LLMs. & \cite{chandran2024private,jacovi2023stop} \\ \cline{2-4}

    & Dynamic Benchmark & Real-time and adaptive evaluation of language models while ensuring data freshness and minimizing contamination risk. & \cite{ma2021dynaboard,li2024latesteval,jain2024livecodebench,fan2024nphardeval} \\ \hline

    \multirow{4}{*}{Data Refactoring} & Data\newline Regeneration & Restructuring and augmenting existing datasets through dynamic evaluation protocols and new prompts. & \cite{xia2024leaderboard,zhu2024dyval,zhu2024dyvala,ying2024seen,yang2023rethinking,wu2024mrke,yu2024freeeval} \\ \cline{2-4}
    
    & Content\newline Filtering & Improving data reliability by identifying and removing contaminated elements. & \cite{dodge2021documenting,ippolito2023preventing} \\ \hline
    
    \multirow{4}{*}{Benchmark-free} & LLM-as-judge & LLMs evaluate themselves without relying on traditional benchmarks. & \cite{li2024treeeval,yu2024freeeval} \\  \cline{2-4}
    
     & Human\newline Participation & Human participant methods involve leveraging human evaluators to assess LLMs performance. & \cite{chiang2024chatbot,yu2024freeeval} \\
    
  \bottomrule
\end{tabular}
\end{table*}

Similarly, \citet{roberts2023data} focused on two code/mathematical problem-solving datasets, Codeforces and Project Euler. They found significant trends that suggest contamination, such as LLM pass rates, correlating with GitHub popularity and release dates of benchmarks. Their work contributes to the field by providing an open-source dataset, raw results, and an evaluation framework, which facilitates further research on BDC. 

We now consider six representative comparison-based studies. Comparison-based strategies offer a robust approach to detecting BDC by scrutinizing model-generation performance against evaluation datasets. These methods, exemplified by various techniques such as content similarity, distribution analysis, perplexity estimation, and temporal performance comparisons, provide valuable insights into the presence and extent of contamination. Comparison-based strategies enable a more comprehensive detection of BDC, offering flexibility in selecting comparison perspectives to identify potential issues. However, akin to matching-based approaches, these methods encounter similar limitations, such as the requirement for substantial computational resources during testing. Additionally, unlike match-based strategies, certain comparison-based strategies may exhibit a restricted scope of detection, concentrating on specific contamination types or datasets. This specificity can hinder generalizability across diverse scenarios; for instance, datasets lacking temporal information impede the application of chronological analysis techniques.

The pursuit of practical solutions for detecting BDC within evaluation datasets is highly important, especially in the context of LLMs. Matching-based methods, focusing on tangible evidence such as dataset inspection and content generation analysis, offer actionable insights into the presence of contamination. However, accessibility to training data and computational demands pose practical challenges to their widespread implementation. Conversely, comparison-based methods provide robust detection mechanisms by scrutinizing model performance against evaluation datasets, offering flexibility in detection perspectives. Nonetheless, these methods may have limited detection scopes and require substantial computational resources. In conclusion, both approaches contribute significantly to understanding and mitigating BDC risks, highlighting the critical need for practical solutions that balance effectiveness with feasibility in real-world applications. Continued research and development in this area are essential for advancing the field of NLP and ensuring the trustworthiness of LLMs.

\section{BDC Mitigation Strategies}
\label{sec:mitigation}

After conducting an extensive survey of research on BDC detection, we now move on to consider the challenge of mitigating BDC. We categorize mitigation strategies into three distinct approaches: \textbf{data curation}, \textbf{data refactoring}, and \textbf{benchmark-free}. These strategies are discussed in Sections \ref{sec:dataCuration}, \ref{sec:dataRefactoring}, and \ref{sec:benchmarkfree}, respectively. We summarize all investigated mitigation strategies in Table \ref{tab:mitigation}.

\subsection{Curating New Data}
\label{sec:dataCuration}

\begin{figure}[h]
  \centering
  \includegraphics[width=100mm]{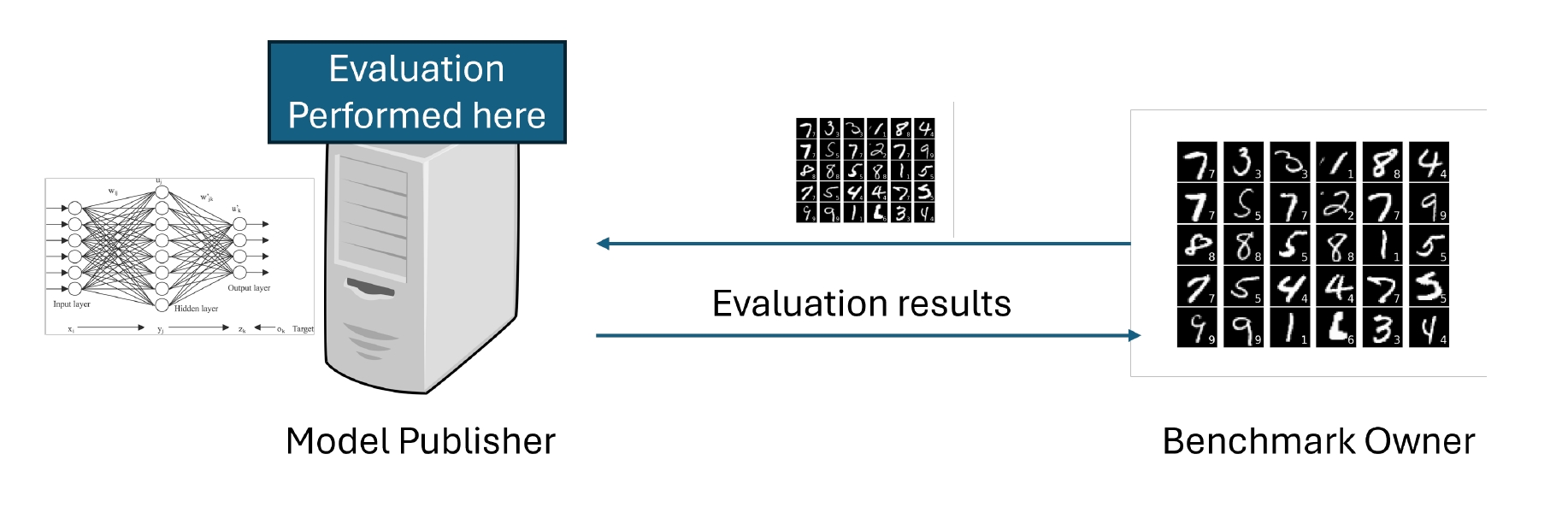}
  \caption{The scheme proposed by \citet{chandran2024private}.}
  \label{fig:chandran2024private}
\end{figure}

Employing new data is the most straightforward way to mitigate the BDC problem \cite{kim2024fables,ghahroodi2024khayyam}. However, this is often an impractical solution. Furthermore, new data is only uncontaminated until it is incorporated into a pre-trained corpus of a future LLM. Addressing the continued availability of new benchmarks is a challenge that has received considerable attention. Along this line of thought, it is easy to associate the use of private datasets to carry out the evaluation of the performance of LLMs so that the benchmarks are less likely to appear in the pre-training data domain. \citet{chandran2024private} proposed a novel approach to benchmarking where test datasets remain private, as shown in Figure \ref{fig:chandran2024private}, preventing contamination and ensuring more accurate evaluations of LLMs. The authors described various scenarios and solutions, including the use of confidential computing and cryptography, to maintain the integrity of benchmarking processes. However, this approach necessitates a degree of trust in both the model provider and the entities responsible for maintaining the benchmark's integrity.

Similarly, \citet{jacovi2023stop} also managed to isolate the evaluation data from the public network, using three practical strategies to mitigate BDC: encrypting test data with a public key, demanding training exclusion controls from API holders, and avoiding data that appears with its solution on the Internet. 

In contrast, \citet{ma2021dynaboard} focused on a different strategy, developing dynamic benchmarks. They introduced Dynaboard, a new platform for evaluating NLP models. Unlike traditional methods that rely on self-reported metrics or predictions on a single dataset, Dynaboard evaluates models directly in the cloud. This approach addresses challenges such as reproducibility and accessibility, allowing for real-time interaction with models and collection of additional metrics like memory use and robustness. Models are ranked on the Dynascore, a utility-based aggregation of these metrics, which can be customized to reflect user preferences.

In a similar vein, \citet{li2024latesteval} introduced LatestEval, an automated method for creating uncontaminated reading comprehension evaluations. LatestEval combats BDC by using texts published within a recent time window, ensuring no overlap with the training corpora of pre-trained language models. The authors developed an automated pipeline to gather the latest texts, identify key information, and construct questions that require models to infer answers from the context rather than copy-pasting. Their experiments showed that language models exhibit minimal memorization behaviors on LatestEval compared to previous benchmarks, suggesting a more robust evaluation and a reduced risk of BDC.

\citet{jain2024livecodebench} described the same idea to the code generation task by proposing LiveCodeBench, a benchmark that evaluates LLMs for coding by using a contamination-free dataset of coding problems from competitive programming platforms. Their results show that LiveCodeBench can effectively measure the generalization capabilities of LLMs and highlight the potential overfitting issues in existing benchmarks. The core of this work is to achieve mitigation of BDC by continuously updating the test cases in the benchmarks. Similarly, \citet{fan2024nphardeval} proposed a dynamic benchmark with monthly updates to test the reasoning ability of LLMs.

Curating new data represents a direct and widely-adopted strategy for mitigating BDC. Within this strategy, we classify the primary methods into two distinct types: \textbf{private benchmark} and \textbf{dynamic benchmark}. The former approach circumvents inclusion in the pre-training dataset of LLMs by isolating newly collected evaluation data from the public network. The key advantage lies in its effective prevention of BDC through straightforward isolation. Encryption and stringent control measures further ensure data integrity. However, the limited accessibility of private benchmarks introduces opacity, necessitating heightened ethical considerations for both model providers and benchmark custodians. On the other hand, dynamic benchmarks offer an intriguing avenue for real-time and adaptive assessment of LLMs. By ensuring data freshness and minimizing contamination risk, they improved model evaluation. Nevertheless, dynamic benchmarks introduce bias and lack the guarantee of consistent results across successive assessments. Note that both approaches require substantial additional resources to facilitate evaluations, potentially constraining their scalability.

\begin{figure}[t]
  \centering
  \includegraphics[width=100mm]{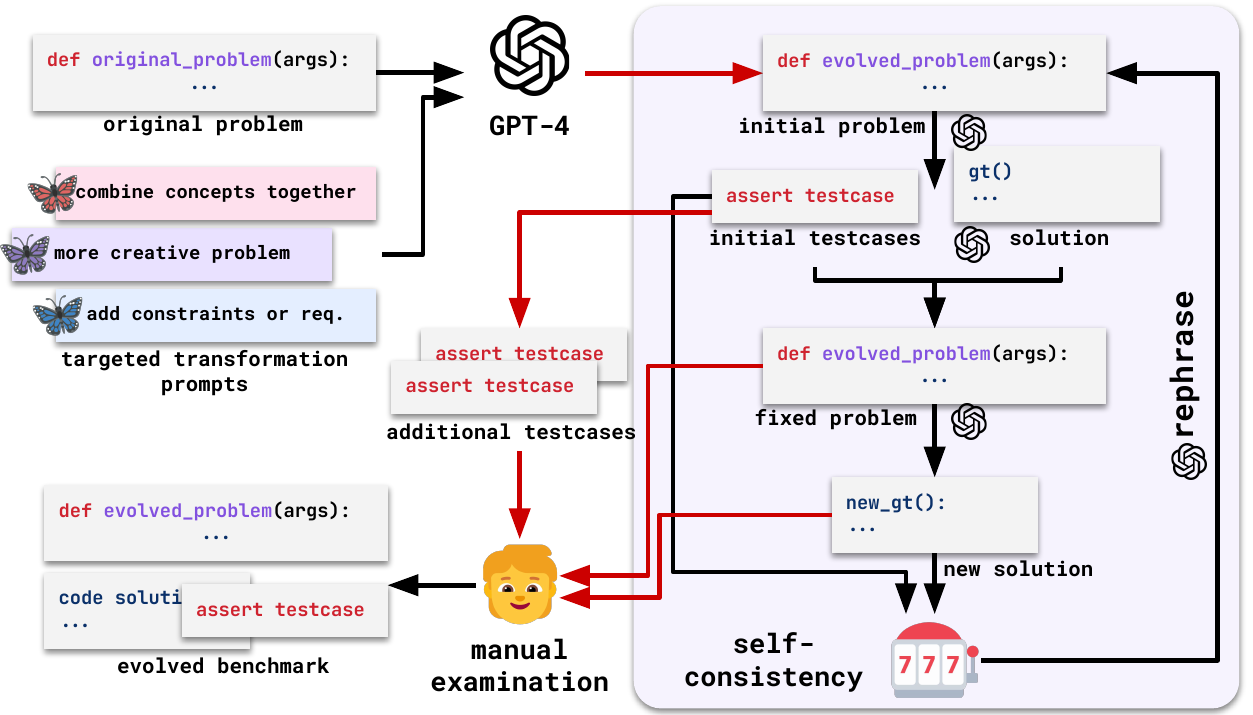}
  \caption{Overview of EVOEVAL evolving problem generation pipeline proposed by \citet{xia2024leaderboard}}
  \label{fig:xia2024leaderboard}
\end{figure}

\subsection{Refactoring Existing Data}
\label{sec:dataRefactoring}

Efforts to address the BDC challenge in LLM evaluations extend beyond curating new data. Strategies now include refactoring existing data, aiming to enhance evaluation reliability and effectiveness by restructuring and augmenting established benchmarks. In this section, we look at innovative methodologies proposed in recent literature, drawing insights from studies such as EvoEval and DyVal 2. These methodologies leverage diverse techniques to refactor existing evaluation datasets. Additionally, content filtering of existing datasets, as demonstrated by \citet{dodge2021documenting}, contributes valuable mitigation against BDC risk.

\citet{xia2024leaderboard} proposed a scheme called EvoEval to mitigate the BDC problem, which focuses on the coding capabilities of LLMs, and which essentially creates five corresponding new prompts based on five dimensions (\textit{Difficult, Creative, Subtle, Combine, Tool use}) for existing test questions, and then uses them to evaluate the consistency of the answers obtained by the model dealing with different prompts about the same question, and the code pass rate to assess its performance. The results show that on 51 LLMs, after applying the component, the models achieved an average of 39.4\% reduction in performance on the HumanEval \cite{chen2021evaluating} benchmark.

\begin{figure}[h]
  \centering
  \includegraphics[width=\linewidth]{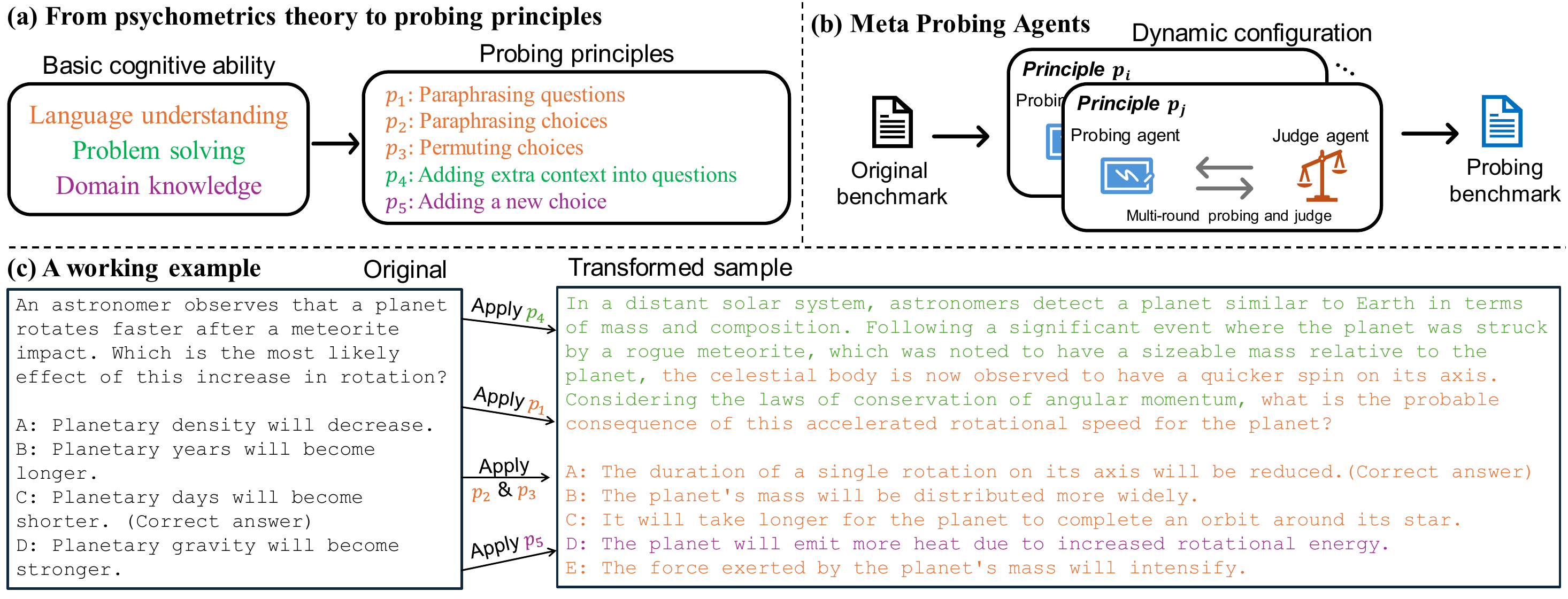}
  \caption{The Meta Probing Agent (MPA) \cite{zhu2024dyval} process that transforms an original benchmark into a new one. The principles here can be combined to create various probing benchmarks for multifaceted analysis. In (c) we see how MPA generates a new sample, given an existing sample from ARC-C~\cite{clark2018think}.}
  \label{fig:zhu2024dyval}
\end{figure}

Similarly, the DyVal 2 \cite{zhu2024dyval} study introduced a new dynamic evaluation protocol called Meta Probing Agents (MPA), which is designed to assess LLMs more effectively. MPA, as a part of DyVal 2, extends the previous DyVal \cite{zhu2024dyvala} framework and focuses on three basic cognitive abilities: \textit{language understanding}, \textit{problem-solving}, and \textit{domain knowledge},  the framework and an example of which are demonstrated in Figure \ref{fig:zhu2024dyval}. The protocol dynamically configures these abilities to provide a multifaceted analysis of LLMs. The extensive evaluations conducted using MPA revealed that most LLMs have room for improvement. The study also found a strong correlation between the basic abilities and model size, indicating that larger models have stronger abilities. Additionally, MPA can serve as a data augmentation method to enhance the capabilities of LLMs.

\begin{figure}[h]
  \centering
  \includegraphics[width=80mm]{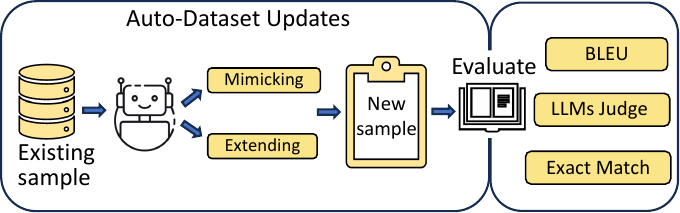}
  \caption{Auto-dataset update framework proposed by \citet{ying2024seen}, who deployed two strategies: mimicking and extending to update.}
  \Description{Auto-dataset update framework}
  \label{fig:ying2024seen}
\end{figure}

\citet{ying2024seen} proposed an innovative approach to maintaining the reliability and timeliness of dataset evaluations for LLMs. In Figure \ref{fig:ying2024seen}, we see that the authors introduced two strategies: a mimicking strategy that uses LLMs to generate new, stylistically similar samples to existing ones, and an extending strategy that adjusts the difficulty of samples based on cognitive levels. Their experiments demonstrated that these strategies can effectively mitigate data leakage issues and provide a more nuanced evaluation of LLM capabilities.

In other work, \citet{yang2023rethinking} proposed a more robust LLM-based decontamination method and apply it to popular pre-training and fine-tuning datasets. They advocate for the adoption of stronger decontamination approaches and the development of fresh one-time exams for accurate model evaluation.

Additionally, \citet{dodge2021documenting} examined the Colossal Clean Crawled Corpus (C4, \citet{raffel2020exploring}), a dataset used to train LLMs. The authors provide a detailed documentation of C4, revealing unexpected sources like patents and US military websites. They also discover machine-generated text and evaluation examples from other NLP datasets within C4. The study evaluates the impact of filters used to create C4, showing that blocklist filtering disproportionately removes text related to minority individuals.

Methods for refactoring existing data, specifically \textbf{data regeneration} and \textbf{content filtering}, represent promising avenues for addressing the challenges posed by BDC in language model evaluations. Data Regeneration, exemplified by approaches such as EvoEval and DyVal 2, emphasizes the restructuring and augmentation of existing datasets to provide multifaceted assessments of LLMs capabilities. By dynamically configuring evaluation protocols and introducing new prompts, these methods enhance the granularity and depth of model evaluations. However, they may require substantial computational resources and expertise to implement effectively. On the other hand, Content Filtering strategies, as demonstrated by the work of \citet{yang2023rethinking}, focus on identifying and mitigating sources of contamination within datasets. These approaches offer more targeted solutions and can provide immediate improvements in data quality. Nonetheless, they may overlook nuanced aspects of model performance and require ongoing adjustments to adapt to evolving challenges. Overall, both Data Regeneration and Content Filtering methodologies contribute valuable insights and tools to the broader endeavor of refining evaluation datasets, underscoring the importance of multifaceted approaches in addressing the issue of BDC in language model evaluations.

 \begin{figure}[t]
  \centering
  \includegraphics[width=\linewidth]{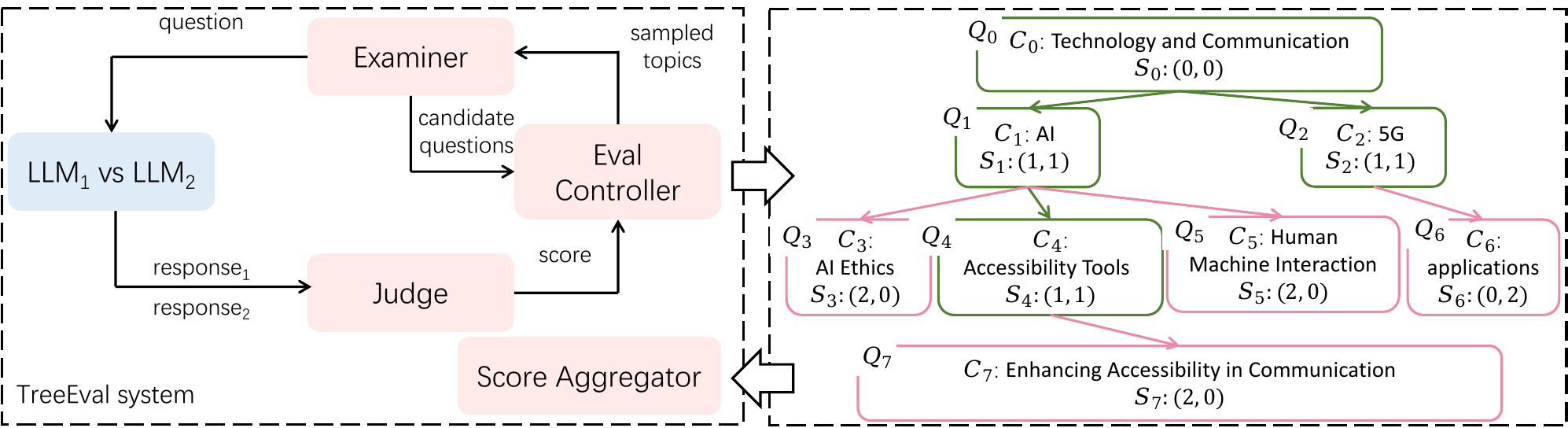}
  \caption{TreeEval \cite{li2024treeeval} system with an illustrative tree for evaluation. The left section contains the components and their workflow in TreeEval. The right section displays a constructed tree within topic Technology and Communication for evaluation (the leaf nodes are shown in red boxes), where each node denotes a question annotated with its topic and valuation score.}
  \label{fig:li2024treeeval}
\end{figure}

\subsection{Benchmark-free Evaluation}
\label{sec:benchmarkfree}

To provide more flexible evaluation methods for LLMs with a reduced risk of contamination, researchers have looked at a more radical strategy: benchmark-free evaluation. This strategy aims to circumvent the BDC risk associated with traditional benchmark assessments. Our review of the current research landscape reveals a nascent yet significant direction within this area. Specifically, we categorize it into two subcategories: \textbf{LLM-as-judge} and \textbf{human participation}. These novel approaches offer promising avenues for addressing the pervasive issue of BDC in LLMs evaluations while fostering greater adaptability and reliability.

LLM-as-judge was first used to measure human preference for content generated by LLMs \cite{zheng2023judging,bai2023benchmarking,wang2023large,zhang2023wider,wang2024pandalm,li2024generative,zhu2024judgelm}, and then \citet{li2024treeeval} suggested that it could be used to mitigate the BDC problem of the LLMs benchmark. They introduced TreeEval, a novel method for evaluating LLMs without relying on traditional benchmarks. It allows a high-performance LLMs to conduct an irreproducible evaluation session, effectively preventing data leakage. As shown in the Figure \ref{fig:li2024treeeval}, this method uses a tree planning strategy to generate a series of questions based on the current evaluation status, ensuring a comprehensive and efficient assessment. The study tested six models of varying sizes and found that TreeEval achieved a high correlation with AlpacaEval2.0 \cite{alpaca_eval,alpaca_eval_length,dubois2023alpacafarm} using approximately 45 questions, demonstrating its robustness and reliability.

Similarly, \citet{chiang2024chatbot} reduced the idea of LLM-as-judge to Human-as-judge, i.e., the use of human participation to evaluate the performance of LLMs while also mitigating the BDC problem, and then they introduced a novel platform called Chatbot Arena\footnote{\url{https://chat.lmsys.org/}} that uses crowdsourced human preferences to evaluate LLMs. It employs a pairwise comparison method and has collected over 240K votes, establishing it as a widely-referenced LLMs leaderboard.

\begin{figure}[h]
  \centering
  \includegraphics[width=100mm]{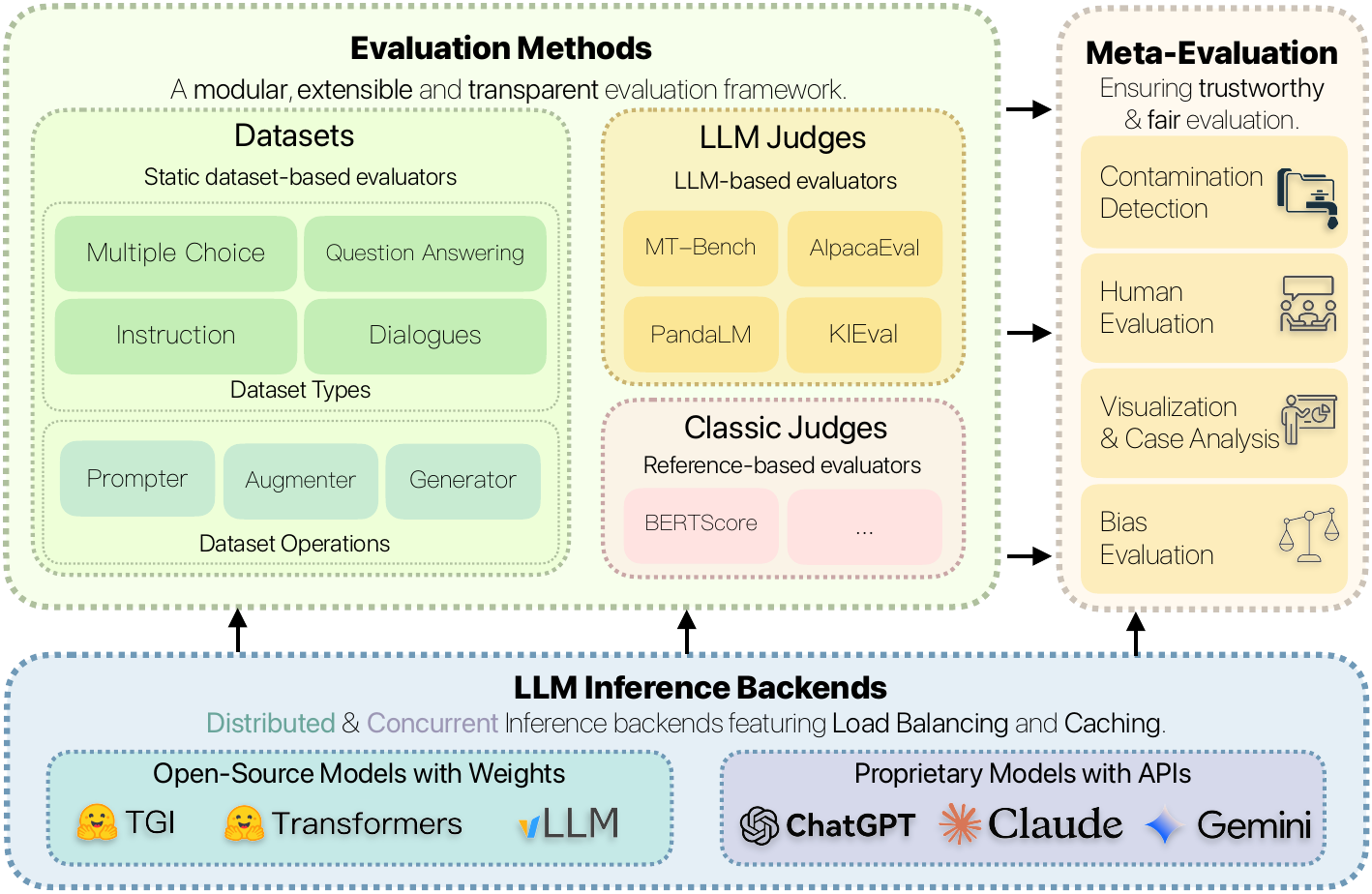}
  \caption{FreeEval framework proposed by \citet{yu2024freeeval}.}
  \Description{FreeEval framework}
  \label{fig:yu2024freeeval}
\end{figure}

Notably, FreeEval proposed by \citet{yu2024freeeval}, integrates multiple methods into one comprehensive framework, including traditional dataset assessment, data regeneration, LLM-as-judge, human participation, and other means which is shown in Figure \ref{fig:yu2024freeeval}. FreeEval is designed to enable trustworthy and efficient automatic evaluations of LLMs, and the key features of FreeEval include unified abstractions for diverse evaluation methodologies, integration of meta-evaluation techniques, and a high-performance infrastructure. 

The emergence of benchmark-free evaluation methods presents a new way of achieving more adaptable and robust language model assessments, particularly in mitigating the risks associated with BDC. In this section, we survey relevant studies and categorize them into LLM-as-judge and human participation. The former minimizes the risk of inadvertently incorporating biased or contaminated data by enabling LLMs to self-evaluate without relying on conventional benchmarks. However, when implementing this evaluation paradigm, careful consideration is necessary regarding whether the LLMs controlling the evaluation process were trained on contaminated or biased pre-training datasets, as this can significantly impact the method's effectiveness. On the other hand, the human participant approach offers valuable insights into LLMs performance within real-world contexts and applications. Nevertheless, human evaluation inherently introduces subjectivity, influenced by personal preferences, potentially leading to discrepancies and biases in the evaluation results. Moreover, collecting and analyzing assessments from human participants is resource-intensive and time-consuming, demanding rigorous efforts to ensure reliability and validity. Both approaches share a common foundation: utilizing external evaluators to test LLMs. However, they diverge in their choice of evaluators—LLMs or humans. Consequently, both approaches face similar challenges, including uncertainty regarding external evaluators and the computational resource demands inherent to this strategy.

We have presented a comprehensive overview of strategies for mitigating BDC in LLM evaluations. We see that these strategies fall into three categories: data curation, data refactoring, and benchmark-free evaluation approaches. Each category offers different solutions to tackle the challenges posed by BDC. Data curation methods, such as private and dynamic benchmarks, focus on isolating evaluation data to prevent contamination, albeit with considerations regarding accessibility and biases. Refactoring existing data through techniques like data regeneration and content filtering enhances evaluation reliability but may require substantial resources. Benchmark-free evaluation, including LLM-as-judge and human participation methodologies, presents radical yet promising alternatives, with each approach offering unique insights into LLM performance while posing challenges related to biases, subjectivity, and resource requirements. Notably, these strategies are not immune to secondary contamination, as newly collected or refactored data may still be influenced by LLMs trained on previously contaminated data. Furthermore, semantic-level contamination remains unavoidable, and simple content filtering may not suffice. Even LLMs acting as evaluators draw on the knowledge of their training data, introducing inherent risks of BDC. Collectively, these multifaceted strategies underscore the complex nature of the BDC challenge and emphasize the need for robust and adaptable LLM evaluations in real-world contexts.

\section{Challenges and Future Directions}
\label{sec:future}

In current research on LLMs and BDC, it is deemed impracticable to fully remove the risks associated with contamination. This is based on two main reasons:
\begin{enumerate}
    \item \textbf{Imperative of Large-Scale Pre-training}: The core of LLMs' capabilities is unlocked through extensive pre-training exercises, which necessitate the use of a substantial level of training data. This content invariably encompasses information pertinent to benchmark datasets. While strategies such as data filtration and regeneration offer some mitigation of BDC risks, they fall short in addressing the semantic and informational dimensions of BDC. Furthermore, any prospective techniques capable of addressing these concerns would likely mandate prohibitive computational resources, rendering them impractical for widespread application.

    \item \textbf{Ascendancy of AIGC}: Coinciding with the increasing maturity and ubiquity of LLM technologies, AI models are increasingly instrumental in generating new content. These models are predicated on large-scale, obscure pre-training datasets. The recursive nature of AIGC enables the evolution of BDC towards semantic dimensions, expanding the seriousness of this problem and making the human identification of BDC risks more challenging.
\end{enumerate}

These factors collectively highlight the complexity of the challenges faced in dealing with BDC risks, underscoring the need for innovative solutions that balance performance with practicality. We outline here several promising future directions for mitigating these problems:
\begin{itemize}
    \item \textbf{Human Evaluation}: The inclusion of human evaluators in evaluating LLMs, as per \citet{chiang2024chatbot}, potentially represents an ideal approach. However, this strategy is not without challenges. Such evaluation processes are resource-intensive and susceptible to various individual background influences, including political affiliations, cultural perspectives, personal beliefs, and educational backgrounds. These contextual factors introduce inherent subjectivity into the evaluation process, potentially leading to unintended biases.
    
    \item \textbf{Dynamic System}: The development of dynamic systems for adaptive evaluation of LLMs is also a promising direction. Adaptive evaluation systems, exemplified by the work of \citet{li2024treeeval}, offer an innovative approach. They operate beyond the confines of traditional benchmark training and testing paradigms, leveraging dynamic scheduling to assess model performance. By doing so, they can reveal the true capabilities of LLMs, thereby mitigating BDC risks to a significant extent and enhancing overall model evaluation. However, a key consideration lies in the data source underpinning the dynamic evaluator. While work such as that of \citet{li2024latesteval} introduces fresh data streams and the evaluation process is adaptive, the data used for evaluation remains inherently static, as well as potentially AIGC data. Consequently, residual BDC risks persist. Therefore, careful scrutiny of the data origin and composition is required to ensure the integrity and reliability of dynamic evaluation systems.
    
    \item \textbf{Benchmark Content Tags}: There have been calls for the establishment of Benchmark Content Tags. Analogous to the robots.txt\footnote{\url{https://www.robotstxt.org/}} protocol employed by search engines or Google's Fact Check Tools API\footnote{\url{https://toolbox.google.com/factcheck/apis}} in the fact-checking field, this protocol aims to improve transparency and facilitate the identification of content associated with benchmark datasets. Specifically, we advocate for the inclusion of standardized tags when posting content relevant to these benchmarks. These fixed tags serve as indicators that model evaluation is implicated. By adopting such a protocol, we mitigate the burden of filtering pre-training data, thereby promoting more effective and efficient model development and evaluation processes.
    
    \item \textbf{Adversarial Evaluation}: The exploration of adversarial evaluation methodologies presents a promising avenue for mitigating the BDC problem in LLMs. This involves the development of generative models, incorporating diverse technological paradigms such as reinforcement learning \cite{puterman2014markov,kaelbling1996reinforcement,suyyon1998reinforcement}, adversarial generative networks \cite{goodfellow2014generative}, and variational autoencoders \cite{kingma2022autoencoding,kingma2014semi}, to synthesize new data representative of natural language while evading potential BDC risks. By harnessing adversarial techniques, these models can generate data that challenges the robustness and generalization capabilities of LLMs, facilitating more rigorous evaluation of model performance in the presence of BDC. Moreover, the integration of a BDC detector within the adversarial evaluation framework enables the supervision and validation of generated data, ensuring its integrity and minimizing the likelihood of BDC contamination.

    \item \textbf{Comprehensive Evaluation Systems}: The concept of a Comprehensive Evaluation System emerges as a natural response to the BDC challenge. Existing mitigation strategies often adopt singular viewpoints, potentially overlooking critical aspects. By integrating multiple perspectives, we can holistically address BDC risks, thereby enhancing evaluation reliability. Frameworks, like the one proposed by \citet{yu2024freeeval}, make it more complicated to assess the performance of LLMs. However, systems that integrate multiple BDC mitigation options, including LLM-as-judge, Human Participation, and other tools, can minimize BDC risks.
    
\end{itemize}

\section{Conclusion}

In this paper, we have explored the complex issue of BDC in LLMs and the wide variety of strategies that have been proposed for mitigating it. We reviewed the detection methods of the BDC problem and categorized them into two classes: Matching-based and Comparison-based methods, each come with their own set of challenges. However, they represent necessary steps towards ensuring the validity of LLM evaluations. Subsequently, we have categorized existing BDC mitigation strategies into three main groups: data curation, data refactoring, and benchmark-free evaluation approaches. Each of these strategies offers unique solutions to the challenges posed by BDC, but none are immune to secondary contamination or semantic-level contamination.

We have also highlighted the challenges and future directions in mitigating BDC risks. The necessity of large-scale pre-training and the ascendancy of AIGC make it nearly impossible to fully eliminate BDC risks. However, several promising future directions have been outlined, including human evaluation, dynamic systems, benchmark content tags, adversarial evaluation, and comprehensive evaluation systems. Each of these approaches has its own set of challenges and considerations, but they all contribute to the ongoing effort to balance performance with practicality in LLM evaluations.

In conclusion, the issue of BDC in LLMs is a multifaceted problem that requires a multifaceted solution. While the strategies and directions discussed in this paper offer promising avenues for mitigating BDC risks, it is clear that more work is needed in this area. As LLMs continue to evolve and become more integrated into our daily lives, the importance of robust and reliable evaluation methods will only increase. We hope that this paper serves as a valuable resource for researchers and practitioners in the field as they navigate the complex landscape of LLM evaluation in the face of BDC.

\section{Author Contributions}

All authors contributed significantly to the conception, design, and execution of this paper. CX played a pivotal role in shaping the core ideas and conceptual framework, leading the research effort and contributing substantially to most section. SG contributed notably to Section \ref{sec:background}, providing valuable context and background information. DG and MTK provided supervision and guidance.  All authors have read and approved the final version of the manuscript.


\bibliographystyle{ACM-Reference-Format}
\bibliography{sample-base}

\end{document}